\documentclass{article}
\usepackage{spconf,amsmath,graphicx}

\usepackage{xspace}

\emergencystretch=\maxdimen
\makeatletter
\DeclareRobustCommand\onedot{\futurelet\@let@token\@onedot}
\def\@onedot{\ifx\@let@token.\else.\null\fi\xspace}

\def\eg{\emph{e.g}\onedot} 
\def\ie{\emph{i.e}\onedot} 
 
\def\etc{\emph{etc}\onedot} 
 
\def\etal{\emph{et al}\onedot}
\makeatother

\title{Exemplar-based Linear Discriminant Analysis for Robust Object Tracking}
\name{Changxin Gao, Feifei Chen, Jin-Gang Yu, Rui Huang, Nong Sang}
\address{Science and Technology on Multi-spectral Information Processing Laboratory,\\
School of Automation, Huazhong University of Science and Technology, Wuhan, 430074, China\\cgao@hust.edu.cn}

\begin{document}
%
\maketitle
\begin{abstract}
Tracking-by-detection has become an attractive tracking technique, which treats tracking as a category detection problem. However, the task in tracking is to search for a specific object, rather than an object category as in detection.  In this paper, we propose a novel tracking framework based on exemplar detector rather than category detector. The proposed tracker is an ensemble of exemplar-based linear discriminant analysis (ELDA) detectors. Each detector is quite specific and discriminative, because it is trained by a single object instance and massive negatives.  To improve its adaptivity, we update both object and background models. Experimental results on several challenging video sequences demonstrate the effectiveness and robustness of our tracking algorithm.

\end{abstract}
\section{INTRODUCTION}
\label{sec:intro}

Visual tracking plays a key role in many computer vision applications, such as surveillance, HCI, video editing, \etc. It has been studied intensively during the past decades~\cite{Wu2013benchmark}, the problem in general still remains challenging due to various factors such as appearance, pose, and scale change of objects, occlusion of objects, illumination variations, cluttered scenes, presence of similar objects, \etc.

Recently, tracking-by-detection method has become an attractive tracking technique~\cite{Avidan2001support, avidan2005ensemble, Babenko2009MIL, collins2005online, hare2011struck}, which treats tracking as a classification problem and trains a detector to separate the object from the background.  Good performance has been shown following this strategy, by borrowing some techniques from object detection methods~\cite{avidan2005ensemble, collins2005online, grabner2006OAB, grabner2008OSB, zeisl2010online, gao2012OTB}.  Furthermore, Stalder~\etal discussed the relationship of tracking, detection, and recognition in \cite{stalder2009BSOBT}.  However, the task in tracking is different from that in detection, that is, finding a specific object instance in tracking, while finding an object category in detection.  Therefore, we suppose that tracking should be based on object exemplar rather than category. That is to say, we should design a specific, and more restrictive detector for the object instance to be tracked, rather than a category-based detector.

To this end, we present an Exemplar-based Linear Discriminant Analysis (ELDA) model for visual tracking. Exemplar-based learner is supposed to be extremely discriminative, because the it is trained by a single object instance and massive amounts of negative samples.  Malisiewicz \etal proposed exemplar-based support vector machine (SVM) algorithm for object detection in \cite{malisiewicz2011ensemble}.  However, training an exemplar-based SVM, by mining for hard negative exemplars, is quite expensive.  Linear Discriminant Analysis (LDA) technique is introduced to improve the speed of training and testing~\cite{hariharan2012discriminative}, which enables exemplar-based method to be used in tracking task.
\begin{figure}[t]
\includegraphics[width=8.5cm]{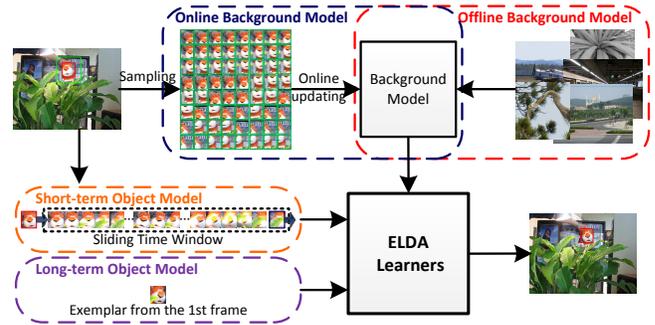}
\caption{An overview of the ELDA tracking algorithm.  ELDA tracker consists of two object models (long-term object model and short-term object model), and two  background models (off-line background model and online background model). The proposed method builds a single ELDA learner for each sample in object model, with both off-line and online background samples. The figure is best viewed in color.
}
\label{fig:framework}
\end{figure}

ELDA algorithm consists of two parts, object model and background model, as in Fig.~\ref{fig:framework}. To achieve good adaptivity of the proposed method, we update both of them.  The object model should give full play to the role of each object exemplar during the tracking process, to handle with the variety of the object.  Although training an ELDA detector is very cheap, taking all the exemplars to build the object model is infeasible, especially in a very long-time tracking.  We only use the exemplars in a predefined sliding time window in this work.  On the other hand, the first frame is very important in tracking, because it includes the precise labels.  Thus, we use this exemplar during the whole tracking process, called long-term object model.  Similarly, we call the previous one short-term object model.

To train a discriminative ELDA detector, vast amounts of negative samples are required. However, we know that, it is difficult and time-consuming to obtain in the procedure of tracking. Thus, we build an initial background model by collecting a sufficient large scale negative set from some natural images with an off-line manner.  On the other hand, the background information just around the object instance is critical for tracking in both discrimination and adaptivity.  Accordingly, we also update background model by an online manner besides off-line one.

To sum up, we present a novel visual tracking framework called ELDA, which is quite discriminative due to exemplar-based learner training by a single object instance and massive negatives, and is quite adaptivity due to online updating both object and background models, as shown in Fig.~\ref{fig:framework}.  We apply this algorithm to visual tracking on several public video sequences and find the results quite promising.

\section{RELATED WORK}
\label{sec:relat}
Prior approaches to object tracking can be roughly divided into two broad categories for build tracking model, namely, generative and discriminative.  We first point the readers to a survey work~\cite{yilmaz2006object} and a recent benchmark work~\cite{Wu2013benchmark}.  Our method is a discriminative tracking model, we therefore present some other previous work on this topic~\cite{Wu2009contextual, kalal2010TLD, kwon2010VTD, santner2010prost, dinh2011CXT, henriques2012CSK, zhang2012CT}. As shown in these works, most discriminative tracking methods are based on haar-like feature representation and online boosting classifier. However, in object detection, the most popular framework is based on HOG feature~\cite{Dalal2005HOG} and linear SVM or LDA. Furthermore, Struck~\cite{hare2011struck}, as one of the most comparative tracker, shows the advantages of SVM classifier in tracking.  Motivated by these works, we introduce this detection framework into tracking problem.  However, the most difference of our method with the state-of-the-art tracking-by-detection methods is that the detectors in our method are trained by an exemplar-based classifiers, rather than category-based classifiers.

Exemplar-SVM detection methods have recently become particularly popular, due to its discrimination ability. Exemplar-based SVM, first introduced in \cite{malisiewicz2011ensemble}, shows good performance by learning an object model from each single object exemplar. However, training an exemplar-SVM, by mining for hard negative exemplars, is quite expensive.  To resolve this problem, Ref.~\cite{hariharan2012discriminative} applies Linear Discriminant Analysis (LDA) technique to speed up. This framework has been widely used in many computer vision applications, \ie, object detection~\cite{malisiewicz2011ensemble, Maji2013part}, image retrieval~\cite{malisiewicz2012exemplar}, mid-level representation discovery for scene classification~\cite{juneja2013blocks}, and action classification~\cite{doersch2013mid}, \etc.
Note that, HOG and exemplar-SVM (or LDA) framework makes effectiveness of these methods. In view of this, we extend exemplar-based LDA method to online tracking case, by updating both object and background models.

\section{ELDA TRACKING METHOD}
\label{sec:method}

In this section, we introduce the ELDA tracking algorithm, and focus on the process of building ELDA detector, and updating both object model and background model.  Fig.~\ref{fig:framework} is an overview of the proposed approach.

\subsection{ELDA detector}
First we present the typical tracking-by-detection algorithms, which train a detector to distinguish a target object from its local background.  Specifically, given a bounding box position $c_k$ (initial position or tracked result position) in frame $k$, a tracker first labels the samples in a predefined training area $R_{t}$, with the size of radius $r$, to $y=1$ and $y=-1$  for positive and negative samples respectively; then trains a classifier using the feature representations of the samples $X$ and corresponding labels $y$; final tracking result is combined by classifying the samples in detection area $R_{d}$ in frame $k+1$.

In ELDA tracking algorithm, we only take the sample exactly at position $c_k$ as positive sample, rather than the samples in a very close area around $c_k$ as in typical tracking-by-detection method.  The representation of the positive sample in frame $k$ is denoted as $X^p_k$.  Then we train an LDA classifier for $X^p_k$, using the covariance matrix $\Sigma_k$ and means $\mu^n_k$ of a negative dataset, which will be introduced in Sec.~\ref{sec:backmodel}.  The final ELDA classifier can be written as:
\begin{equation}
H_k(X) = sign(\omega_k^T \times X + b_k)
\label{eqn:ELDAclassifier}
\end{equation}
where $\omega_0$ is the threshold, the weights can be calculate by:
\begin{equation}
\omega_k = \Sigma^{-1} ( {X^p_k} - \mu^n_k )
\label{eqn:ELDAweights}
\end{equation}

\subsection{Object Model}
We build object model for each positive exemplar, thus, the key is how to choose the positive samples.  The first frame with the precise labels is critical in tracking.  Thus, we use the object exemplar in the first frame during the whole tracking process, called long-term object model, denoted as $H_1(X)$.   On the other hand, to improve the adaptivity to the variety of object appearance, we would better build object model by applying as many as possible ELDA detectors in theory. However, it's not a good solution in practical application, especially for long-time tracking. In this paper, we simply set a time window $TM$ to choose positives.  Only the samples in past $TM$ frames from frame $k$ are used to build object model, namely, $X^p_i$, $i\in[\begin{array}{*{20}{c}} l & k  \end{array}]$,  $l=max(2, k-TM+1)$, and we call it short-term object model.  The weights of the ELDA detectors $H_k(X)$ are determined by a semi-supervised way using the long-term detector $H_1(X)$ as the prior, then set the weight $\lambda_k$ to $H_k(X)$ as follows:
\begin{equation}
{\lambda_k } = \frac{H_1(X^p_k)}{H_1(X^p_1)}
\label{eqn:semi_weight}
\end{equation}

Accordingly, the object model can be defined as:
\begin{equation}
M^{O}_{k}(X) = \lambda_1 H_1(X) + \sum\limits_{i} {\lambda_i H_i(X)}
\label{eqn:Objectmodel}
\vspace{-0.2cm}
\end{equation}

\subsection{Background Model}
\label{sec:backmodel}
In exemplar-based framework, the background model $M^{B}_{k}$ can be denoted as $(\Sigma_k, \mu^n_k)$ according to Eq.~\ref{eqn:ELDAclassifier} and Eq.~\ref{eqn:ELDAweights}. In the tracking case, the negative samples in a ring area centered at the object position are critical, where we sample online negatives as most of other tracking-by-detection approaches~\cite{grabner2006OAB, Babenko2009MIL}.  On the other hand, the huge number of negatives are the guarantee of discrimination ability of ELDA.  However, it is difficult and time-consuming to obtain lots of negatives in the procedure of tracking.  Therefore, we utilize a strategy to build a background model with large scale negatives, collecting by both off-line and online manner.  To build the off-line model, we first collect massive amounts of negative samples from some natural images, and then calculate the background model $(\Sigma_0, \mu_0)$ as initial model $M^{B}_{0} = (\Sigma_0, \mu_0)$, $X_0$ is the representations of all negative samples.

Online background model is used to improve the adaptivity by some negative samples quite relevant to the tracking task,  that is in the ring area mentioned above.  In frame $k$, we calculate online model $M^{Bon}_k$ using the negative samples $X^n_k$.  The final background model in frame $k$ is incrementally calculated using $M^{B}_{k-1}$ and $M^{Bon}_k$ according to definition of covariance matrix and means.

\section{EXPERIMENTAL RESULTS}
\label{sec:exprel}

\subsection{Experimental setup}
In this section, we evaluate our ELDA tracker on ten public available benchmark video sequences used in previous works~\cite{Babenko2009MIL, hare2011struck, Wu2013benchmark}, namely, david indoor (david), sylvester, singer2, coke can (coke), girl, david outdoor (david3), suv, liquor, woman, and tiger1.  These videos are very difficult to track, because of various challenges, such as, occlusion of objects, illumination variation, appearance change of objects, rotation and scale change of objects, clutter scenes, presence of similar objects, \etc. Some samples corresponding to these challenges can be seen in Fig.~\ref{fig:resultsBbox}, and more details are introduced in \cite{Wu2013benchmark}. All the settings of videos are same as in \cite{Wu2013benchmark}, \eg, tiger1 starts from frame 6.

In our experiments, ELDA tracker is compared with six state-of-the-art tracking-by-detection algorithms, the fragment tracker (Frag)~\cite{adam2006Frag}, the online boosting tracker (OAB)~\cite{grabner2006OAB}, the visual tracking decomposition algorithm (VTD)~\cite{kwon2010VTD}, the multiple instance learning tracker (MIL)~\cite{Babenko2009MIL}, the incremental visual tracking method (IVT)~\cite{Ross2008IVT}, and Struck~\cite{hare2011struck}.

\subsection{Implementation details}
For the representation, we use HOG feature ( $8\times 8$ cells with 9 orientations ) in this work. Thus, the resulting feature is $8\times 8\times 4\times 9 = 2304$ dimensions.  To build short-term object model, we set the size of time window $TM=500frames$, which is determined experimentally. To build off-line background model, we collected more than 1000, 000 patches ($64\times 64$ pixels) by randomly sampling on the 5096 images of PASCAL VOC 2008 dataset~\cite{pascal-voc-2008}. Then HOG feature is extracted to build initial background model $M^B_0$; and the online negatives are sampled in the ring area with $5<d\leq30$.  The detect area $R_d$ is also set to $30$.

\subsection{Quantitative Comparison}

\begin{figure*}[!ht]
\centering
\includegraphics[width=0.99\linewidth]{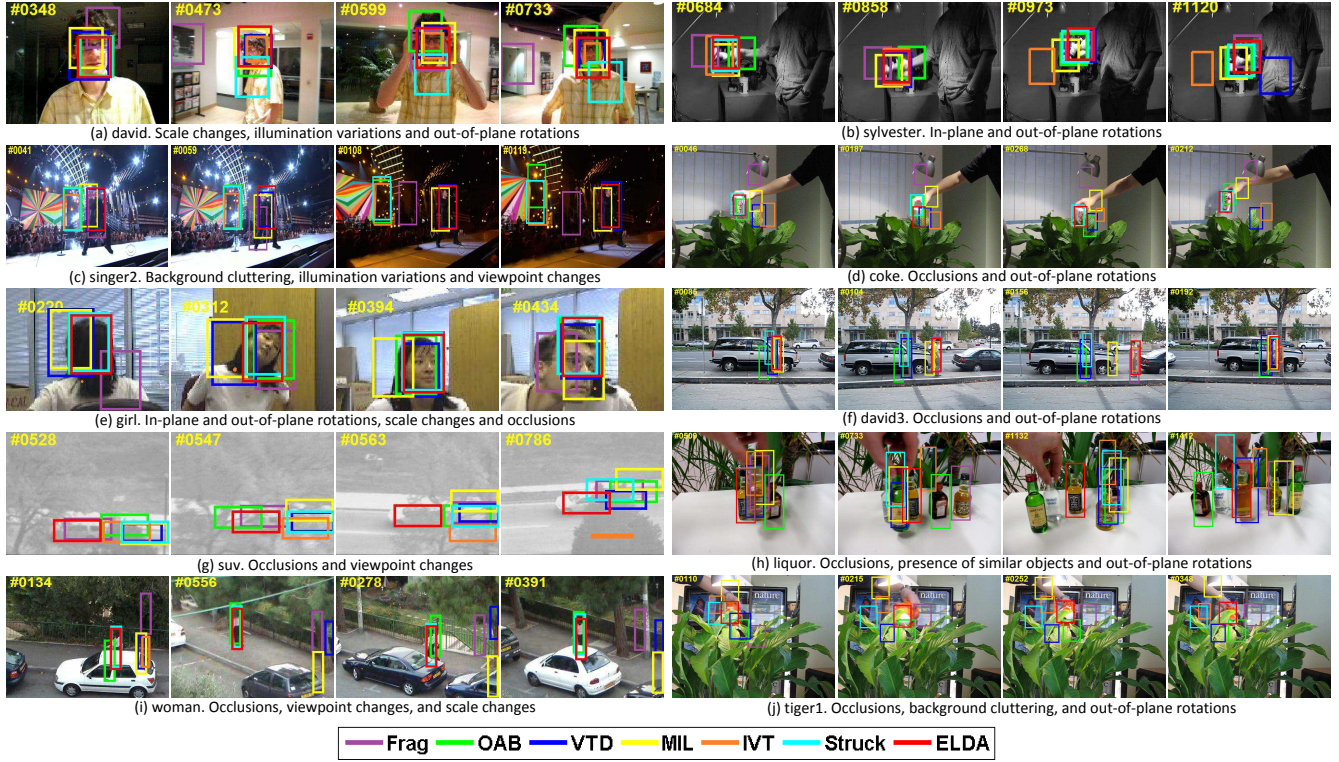}
\vspace{-0.4cm}
\caption{\label{fig:resultsBbox} Tracked bounding box results comparisons of 7 trackers in 10 videos under some special challenges. }
\vspace{-0.5cm}
\end{figure*}

Two common evaluation criteria are used for quantitative comparison, namely, center location error ($CLE$) and success rate ($SR$). First we define these two criteria briefly. For each frame, the result is denoted as tracked bounding box $B_T$ and center location $C_T$, which of ground truth is $B_G$ and $C_G$ respectively. $CLE$ is defined as the average Euclidean distance (in pixels) between $C_T$ and $C_G$. $SR$ is defined as the rate of successful frames in total frames. A tracked result is considered to be successful if the overlap ratio $\frac{area(B_T \bigcap B_G)}{area(B_T \bigcup B_G)}$ is larger than 0.5.

Table~\ref{tableCLE} and Table~\ref{tableSR} report the comparison results of ELDA and other six state-of-the-art trackers in terms of average center location error and success rate.  It can be seen that, ELDA tracker outperforms other trackers on 5 out of 10 videos, and obtains 8 best or second best scores out of 10 videos in terms of both average center location error and success rate. Most exiting, ELDA tracker, over all, performs well against other six state-of-the-art algorithms.  Note that only one average center location error is bigger than 20 pixels in our results, which demonstrate the proposed method works robustly.

\begin{figure*}[tbp]
\centering
\includegraphics[width=0.99\textwidth]{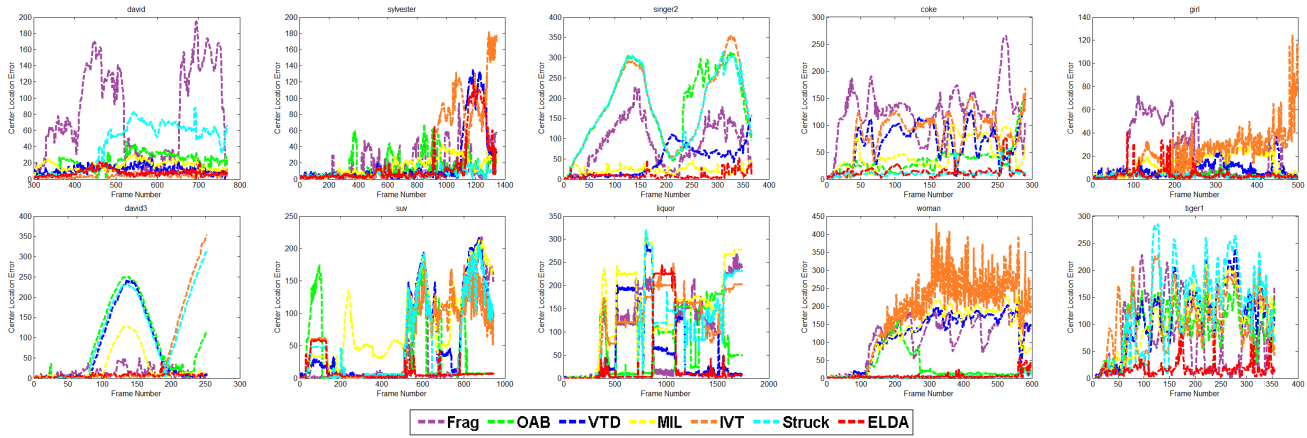}
\caption{\label{fig:results} Tracking error (in pixels) of 7 trackers on 10 video sequences. The figure is best viewed in color.}
\end{figure*}

To highlight the superior performance of the ELDA tracker, we show some images with comparison tracked bounding box in Fig.~\ref{fig:resultsBbox}  under lots of special challenges, e.g., heavy occlusion, illumination variations, appearance changes, rotation and scale changes, background cluttering, presence of similar objects occur. To present the tracking results frame by frame, we also give the corresponding tracking error of 7 trackers on 10 video sequences in Fig.~\ref{fig:results}.  It shows the good performance of ELDA tracker in both accuracy and adaptivity.

\begin{table}[!t]
\caption{Average center location error (in pixels) comparison of the 7 trackers on 10 videos. Bold and underlined values indicates best and second best performance.}
\label{tableCLE}
\vspace{-0.5cm}
\begin{center}
\small
\begin{tabular}{p{31pt}p{17pt} p{17pt} p{17pt} p{17pt}p{17pt} p{17pt}p{17pt}} 
\hline
Sequence&Frag   &OAB    &VTD    &MIL    &IVT    &Struck &ELDA\\ \hline
david	&82.1 	&21.7 	&11.6 	&16.9 	&\textbf{4.8} 	&42.8 	&\underline{7.9}\\
sylvester	    &15.0 	&\underline{14.8} 	&19.6 	&15.2 	&34.2 	&\textbf{6.3} 	&17.3\\
singer2	&88.6 	&185.9 	&43.7 	&\underline{22.5} 	&175.5 	&174.3 	&\textbf{9.3}\\
coke	&124.8 	&35.9 	&68.6 	&46.7 	&83.0 	&\textbf{12.1} 	&\underline{14.3}\\
girl	&20.7 	&\underline{3.7} 	&8.6 	&13.7 	&22.5 	&\textbf{2.6} 	&3.7\\
david3	&\underline{13.6} 	&83.4 	&66.7 	&29.7 	&51.9 	&106.5 	&\textbf{6.8}\\
suv	    &42.0 	&\underline{30.5} 	&57.2 	&82.2 	&57.3 	&49.8 	&\textbf{9.7}\\
liquor	&99.6 	&68.6 	&\underline{60.2} 	&141.9 	&118.5 	&91.0 	&\textbf{33.2}\\
woman	&111.9 	&31.4 	&118.9 	&125.3 	&176.5 	&\textbf{4.2} 	&\underline{5.6}\\
tiger1	&\underline{74.3} 	&94.9 	&107.3 	&108.9 	&106.6 	&128.4 	&\textbf{16.9}\\
MEAN	&67.3 	&57.1 	&\underline{56.2} 	&60.3 	&83.1 	&61.8 	&\textbf{12.5}\\ \hline
\end{tabular}
\end{center}
\vspace{-0.5cm}
\end{table}

\begin{table}[!t]
\caption{Success rate comparison of the 7 trackers on 10 videos. Bold and underlined values indicates best and second best performance.}
\label{tableSR}
\vspace{-0.5cm}
\begin{center}
\small
\begin{tabular}{p{31pt}p{17pt} p{17pt} p{17pt} p{17pt}p{17pt} p{17pt}p{17pt}} %
\hline
Sequence&Frag   &OAB    &VTD    &MIL    &IVT    &Struck &ELDA\\ \hline
david	&0.12 	&0.15 	&\underline{0.68} 	&0.23 	&\textbf{0.79} 	&0.24 	&0.61\\
sylvester	    &0.68 	&0.68 	&0.80 	&0.55 	&0.68 	&\textbf{0.93} 	&0.79\\
singer2	&0.20 	&0.09 	&0.03 	&0.48 	&0.04 	&0.04 	&\textbf{0.94}\\
coke	&0.03 	&0.11 	&0.17 	&0.12 	&0.13 	&\textbf{0.94} 	&\underline{0.66}\\
girl	&0.54 	&0.46 	&\underline{0.94} 	&0.29 	&0.19 	&\textbf{0.98} 	&\underline{0.94}\\
david3	&\underline{0.81} 	&0.34 	&0.48 	&0.68 	&0.63 	&0.34 	&\textbf{0.99}\\
suv	    &0.71 	&\underline{0.76} 	&0.55 	&0.13 	&0.44 	&0.57 	&\textbf{0.87}\\
liquor	&0.37 	&0.48 	&\underline{0.58} 	&0.20 	&0.21 	&0.41 	&\textbf{0.85}\\
woman	&0.18 	&\underline{0.61} 	&0.18 	&0.19 	&0.18 	&\textbf{0.93} 	&\textbf{0.93}\\
tiger1	&\underline{0.31} 	&\underline{0.09} 	&0.12 	&0.10 	&0.08 	&0.18 	&\textbf{0.83}\\ MEAN	&0.40 	 &0.42 	&0.46 	&0.30 	&0.34 	&\underline{0.56} 	&\textbf{0.84}\\ \hline
\end{tabular}
\end{center}
\vspace{-0.5cm}
\end{table}

\section{CONCLUSIONS AND FUTURE WORK}
\label{sec:conclu}

The task in tracking is to search for a specific object instance, rather than an object category as in detection. In view of this, we proposed a new tracking framework based on exemplar detector rather than category detector. We build ELDA tracker by both updating object and background models.  Promising results on challenging video sequences demonstrate that our method outperforms the state-of-the-art tracking algorithms. We are considering the following for the future work.  First, in our current tracker, updating strategy of object model with a predefined time widow is very simple. To further improve the adaptivity, we are looking for a more effective updating method.  Second, due to the successful of part-based model in object detection~\cite{Felzenszwalb2010DPM}, we will study part-based tracking approach, to deal with some challenges, \eg, the occlusion, deformation.

\section*{Acknowledgments}
This work is supported by the Project of the National Natural Science Foundation of China No.61271328, and the Fundamental Research Funds for the Central Universities No. HUST2013TS132.

{
\bibliographystyle{IEEEbib}
\bibliography{refs}
}

\end{document}